\pdfoutput=1

\documentclass[11pt]{article}

\usepackage[final]{acl}

\usepackage{times}
\usepackage{latexsym}

\usepackage{booktabs}
\usepackage{multirow} 
\usepackage{makecell}

\usepackage{tcolorbox}
\tcbuselibrary{breakable}  
\usepackage{amsmath}

\usepackage[edges]{forest}
\usepackage{tikz}
\usepackage[utf8]{inputenc}
\usepackage{geometry}

\usepackage[T1]{fontenc}

\usepackage[utf8]{inputenc}

\usepackage{microtype}

\usepackage{inconsolata}

\usepackage{graphicx}

\title{How Far Can LLMs Improve from Experience? Measuring Test-Time Learning Ability in LLMs with Human Comparison}

\author{Jiayin Wang, Zhiqiang Guo, Weizhi Ma, Min Zhang \\
        Tsinghua University, Beijing, China\\
        JiayinWangTHU@gmail.com}

\begin{document}
\maketitle
\begin{abstract}
As evaluation designs of large language models may shape our trajectory toward artificial general intelligence, comprehensive and forward-looking assessment is essential. Existing benchmarks primarily assess static knowledge, while intelligence also entails the ability to rapidly learn from experience. To this end, we advocate for the evaluation of \textbf{Test-time Learning}, the capacity to improve performance in \textit{experience-based, reasoning-intensive tasks} during test time.
In this work, we propose semantic games as effective testbeds for evaluating test-time learning, due to their resistance to saturation and inherent demand for strategic reasoning. We introduce an objective evaluation framework that compares model performance under both limited and cumulative experience settings, and contains four forms of experience representation.
To provide a comparative baseline, we recruit eight human participants to complete the same task. Results show that LLMs exhibit measurable test-time learning capabilities; however, their improvements are less stable under cumulative experience and progress more slowly than those observed in humans.
These findings underscore the potential of LLMs as general-purpose learning machines, while also revealing a substantial intellectual gap between models and humans, irrespective of how well LLMs perform on static benchmarks.
Code and data are available\footnote{https://github.com/Alice1998/Test-time-Leaning}.

\textit{"Give a man a fish, and you feed him for a day; teach a man to fish, and you feed him for a lifetime."}
\end{abstract}

\section{Introduction}

\begin{figure}[htbp]
    \centering
\includegraphics[width=\linewidth]{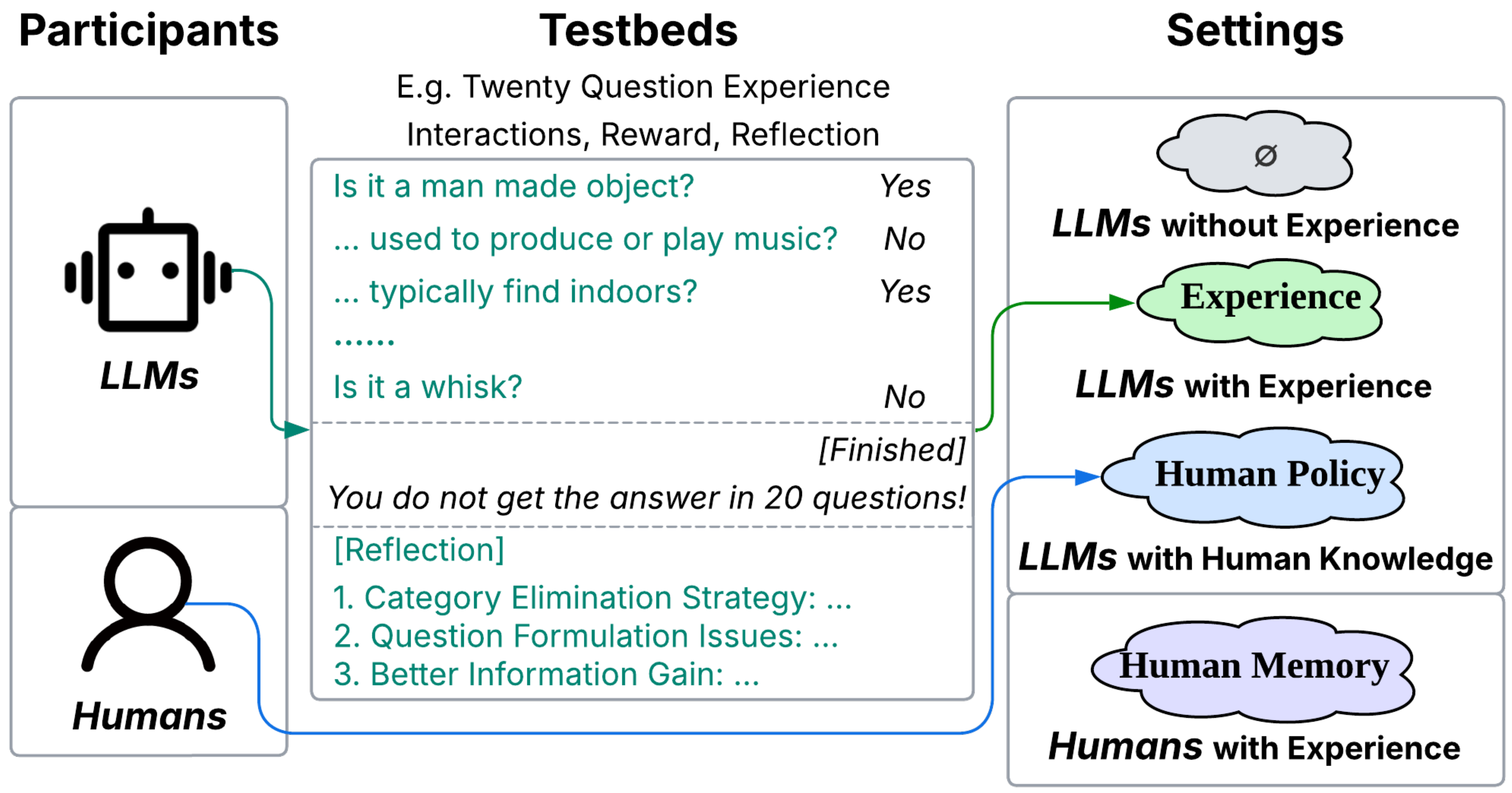}
    \caption{Test-time Learning Evaluation Pipeline.}
    \label{figure:introduction}
\end{figure}

As large language models continue to advance, the design of their evaluations becomes increasingly important, as it shapes the development priorities of the next generation of models and guides the broader trajectory toward artificial general intelligence~\cite{chang2024survey}.
Current benchmarks predominantly focus on measuring the expertise of language models in performing specific tasks. However, intelligence is not solely defined by the possession of expert knowledge~\cite{krathwohl2002revision, minsky1988society}. 
For example, individuals without profound knowledge can still demonstrate intelligence through their speed to acquire new skills through experience~\cite{silver2025welcome}. This dimension of intelligence, the capacity for rapid learning, remains largely overlooked in existing evaluation frameworks.

Assessing the ability to learn quickly is challenging. Under the current LLM development paradigm, models undergo massive pre-training followed by domain-specific alignment~\cite{achiam2023gpt,liu2024deepseek}. Models are typically compared based on their final performance outcomes, without constraints on the amount of task-relevant training data they utilize. In this paper, we do not argue for altering this training paradigm, as it is the reason of success. Rather, we aim to design an evaluation framework that can directly measure a model's ability to improve rapidly at test time.

The characteristic we aim to evaluate aligns closely with the concept of ``Test-time Learning'', which refers to a model’s ability to adapt and improve through its own test-time experience. The desirable end-state for artificial intelligence should have the ability to effectively improve its performance through a limited number of experiences by interacting with environments, reflecting on feedback and rewards, rapidly acquiring in-context or in-weight policies, and acting adaptively. Moreover, these test-time improvements should be capable of accumulating as experience grows, enabling continual adaptation and learning. 

The concept of \textit{test-time learning} shares conceptual similarities with \textit{in-context reinforcement learning}~\cite{laskin2022context,lee2023supervised,grigsby2023amago,lu2023structured} and \textit{agent self-evolution}~\cite{tao2024survey} to some extent. However, it is distinguished from these paradigms in several fundamental ways.
(1) Generality of Environment: In-context reinforcement learning typically operates within classical RL domains such as adversarial bandits~\cite{duan2016rl} or the dark room setting~\cite{laskin2022context}, which are characterized by constrained environments and limited action spaces. In contrast, test-time learning emphasizes generalization in open-ended environments, where the action space spans the full token space of a language model. (2) Beyond Memorization: Research on agent self-evolution has largely focused on tool-use or coding tasks that rely heavily on rote memorization or repeated exposure to similar instances~\cite{qian2024investigate}. These setups often allow models to improve simply by recalling prior examples. Test-time learning is achieved in experience-based, reasoning-intensive tasks that require discovering latent patterns and executing self-proposed policies beyond surface-level recall.

In this work, we propose an \textbf{objective framework to evaluate the test-time learning ability} of current large language models. Rather than relying on static tasks like academic olympiads, we adopt competitive games, which are dynamic, resistant to saturation, and embed latent strategies, making them ideal for studying test-time learning.

Furthermore, we systematically evaluate performance across four test-time experience settings: (1) full experience with interactions, rewards, and model reflection; (2) model-derived policy based solely on game rules; (3) model-derived policy informed by both rules and accumulated test-time experience; and (4) human-authored policy.
To compare LLM performance with human reasoning, we also recruit human annotators to perform the same task. The results reveal clear gaps between human and model test-time learning capabilities, highlighting promising directions for future research.

The experiment results show that LLMs demonstrate measurable test-time learning ability; however, these gains are not stable and consistent when experience accumulates. In contrast, human participants exhibit more stable and rapid learning. These findings highlight the need for further evaluation and improved training strategies to enhance the test-time learning of LLMs.

Importantly, our aim is not to build an elaborate framework centered on agentic workflows, but rather to propose a lightweight and objective pipeline for assessing whether models can benefit from test-time experience. We believe that systematic evaluation of test-time learning constitutes a key step toward advancing the capabilities of large language models.




\section{Related Work}

\subsection{Test-time Learning}

The concept "test-time learning" shares certain similarity with "test-time training", "in-context reinforcement learning" and "self-evolution" but also adopts key distinctions in its focus and formulation.

The first two concepts involve weight updates.
Test-time training~\cite{sun2020test,liu2021ttt++,gandelsman2022test,sinha2023test,sun2019test} primarily addresses distributional or domain shifts between training and test data by adapting model parameters at inference time.
In-context reinforcement studies~\cite{laskin2022context,lee2023supervised,grigsby2023amago,lu2023structured} involves training models from scratch to perform reinforcement learning tasks via in-context tokens.
Self-evolution studies focus on performance improvements through in-context interactions without parameter updates~\cite{tao2024survey, lu2023self}. For instance, \citet{lange2024large} proposed prompting strategies that enhance performance through structured interactions, and \citet{yu2025policyevol} introduced agentic workflows that integrate human knowledge to guide model behavior and maximize gains.
Most prior work emphasizes engineering pipelines to improve in-context performance, often in application settings such as web navigation, tool use, or code generation—domains where improvements are frequently driven by retrieval or surface-level similarity to prior examples, rather than by the development of general strategies or deeper reasoning. As noted by \citet{silver2025welcome}, “now is the time of experience,” highlighting the emerging view that future intelligent agents must learn through interaction to achieve higher-level of reasoning, rather than rely solely on static question answering-style evaluations.


In this work, we aim to objectively evaluate the extent to which LLMs can leverage experience at test time. Specifically, we quantify models' test-time gains and compare them against improvements guided by human-authored policies and human learning trajectories.

\subsection{Evaluation Environments}

For reasoning-intensive evaluation environments, in-context reinforcement learning studies have explored semantic and visual representations of reinforcement tasks such as the adversarial bandit~\cite{laskin2022context,lee2023supervised}, dark room~\cite{laskin2022context,lee2023supervised} and Partially Observable Process Gym~\cite{morad2023popgym,lu2023structured}, a set of simple environments designed to benchmark memory in deep RL. However, these tasks involve closed-ended environments with limited action spaces and are often easily solved by current large language models, as they may already encode effective policies, e.g., upper confidence bound\cite{garivier2011upper}. In contrast, we focus on open-ended experience-based reasoning-intensive tasks with token-level action spaces and moderate difficulty, where the optimal policy is not readily accessible or encoded in the model.
Regarding the self-evaluation of LLMs, recent works have employed web-used~\cite{yao2022webshop}, tool-assisted~\cite{lu2023self}, or static benchmarks, including math~\cite{cobbe2021gsm8k}, code generation~\cite{jiang2023selfevolve,luo2023wizardcoder}, and general-purpose benchmarks~\cite{vicuna2023}. However, these static evaluations are prone to saturation, and observed improvements may result from memorization or recall rather than from enhanced reasoning via learned policies.

In this work, we propose competitive game environments as effective testbeds for evaluating the test-time learning ability of LLMs. These environments are dynamic, resistant to saturation, open-ended, reasoning-intensive, and policy-driven, making them well-suited for assessing model ability to learn and adapt through experience.

\section{Test-time Learning}

\subsection{Testbeds}

The optimal environment for evaluating the test-time learning ability of large language model should satisfy the following requirements: 1) Moderate Difficulty: The environment should not admit a readily accessible optimal policy, either due to the nature of the task or the current limitations of large language models.
2) Structured Regularity: Tasks should contain underlying patterns that can be uncovered and leveraged through interaction and reasoning to enhance performance.
3) Beyond Memorization: Success should depend not on recalling previous answers, but on identifying generalizable rules or strategies that drive improvement.


These criteria highlight the importance of reasoning over purely knowledge-rich contexts. Classic reinforcement learning settings, such as adversarial bandit~\cite{laskin2022context}, have been rendered less meaningful for test-time learning evaluations, as many models have already internalized algorithms like Upper Confidence Bound in their knowledge. To address this, we adopt three diverse environments to evaluate test-time learning: a mathematics benchmark, a single-agent semantic game, and a multi-agent semantic game.


\noindent
\textbf{AIME 2025}~\cite{AIME} refers to the American Invitational Mathematics Examination 2025, used to identify candidates for the U.S. team in the International Mathematical Olympiad (IMO). We leverage this most recent mathematics benchmark to examine the test-time learning capabilities of large language models in solving high-level mathematical problems.

\noindent
\textbf{Twenty Question}~\cite{abdulhai2023lmrl,zhou2024archer} is a dialogue-based multi-turn single agent task in which a large language model attempts to identify a target word from a fixed set of 157 candidate words by asking up to twenty yes/no questions. The environment responds with "Yes", "No", or "Invalid" if the question is not a valid yes/no query. To ensure consistent understanding of the questions, the environment is simulated using the same LLM as the questioning model.

The 157 candidate words, adopted from prior work~\cite{zhou2024archer}, span diverse categories including animals, art, clothes, electronics, fruits, furniture, garden supplies, jewelry, kitchen tools, musical instruments, nature, office supplies, sports, tools, toys, vegetables, and vehicles. The candidate set remains fixed across the games, providing a controlled setting to evaluate whether the LLM can learn effective categorization and formulate increasingly informative dichotomous questions during test time.
Performance is measured by NDCG@20 based on the rank of the correct guess.

\noindent
\textbf{Who is undercover}~\cite{xu2023magic} is a dialogue-based multi-turn multi-agent task. Each player is assigned a secret word: one player receives a distinct word as the undercover, while all others, civilians, share the same word. In each round, players provide verbal clues related to their secret words. By analyzing both their own and others’ clues, players attempt to infer their roles. The objective for civilians is to identify the undercover, while the undercover aims to conceal their identity.
Note we use the neutral word "difference" and "normal" instead of "undercover" and "civilian" in task instructions. This is motivated by the observation that large language models often refuse to acknowledge being the "undercover" due to value misalignment, as further discussed in Appendix~\ref{section:value}.
The performance is evaluated based on the win rate.

\subsection{Test-time Learning Setting}


\begin{figure*}[htbp]
  \includegraphics[width=\linewidth]{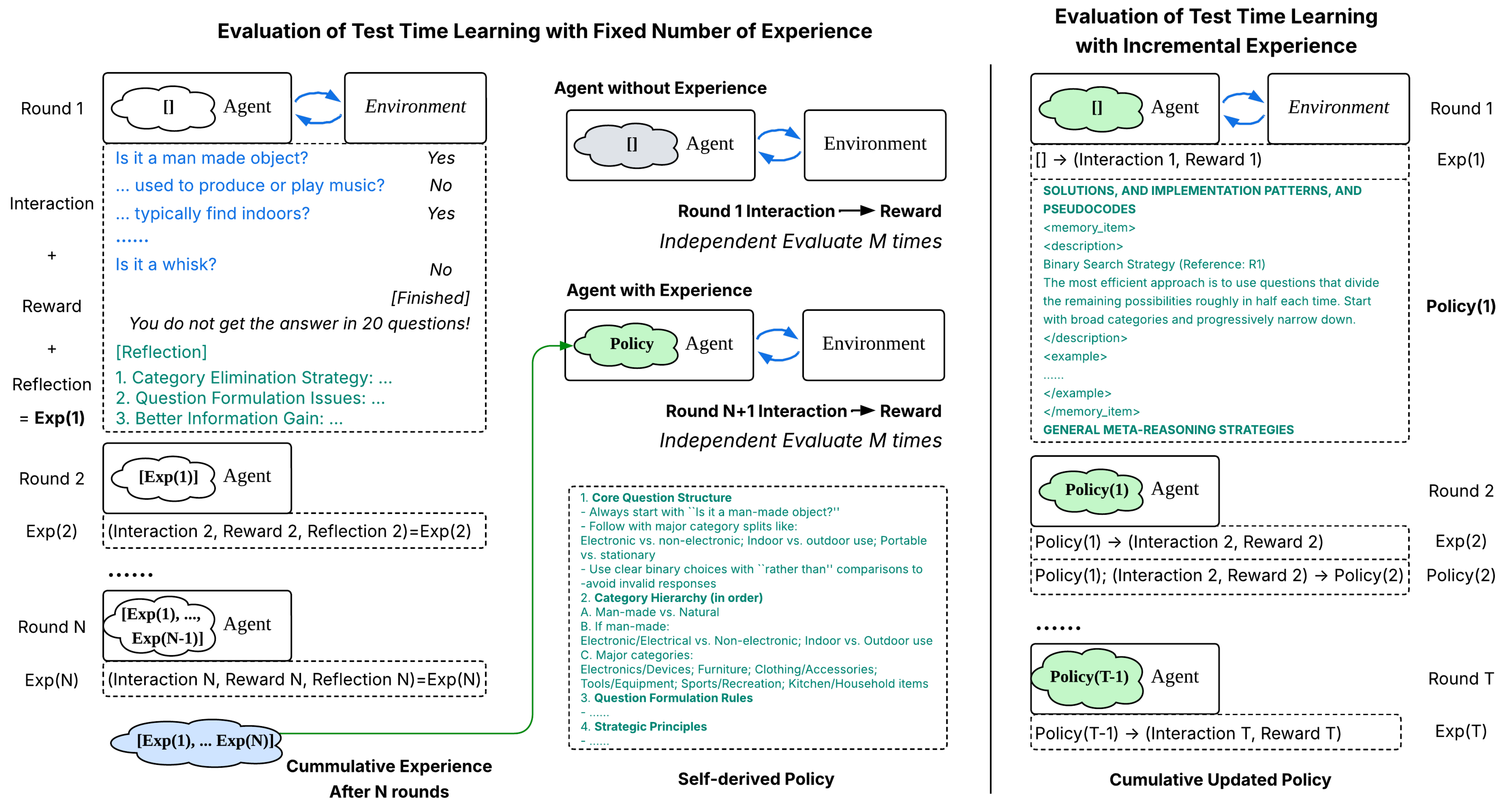} 
  \caption {Test-time Learning Evaluation Settings with Fixed Amount and Incremental Experience.}
  \label{Figure:framework}
\end{figure*}


It is important to note that our objective is not to design an elaborate framework for maximizing task completion rates. Rather, we aim to provide a lightweight and objective evaluation framework that assesses a model’s test-time learning, comparing its performance with and without prior experience, as well as against human-authored policies grounded in human reasoning.
To this end, we adopt a vanilla evaluation setup consisting of two settings: a fixed number of experience setting~\cite{laskin2022context} and an incremental experience setting~\cite{suzgun2025dynamic}.

\subsubsection{Evaluation with Experience}

\begin{table}[htpb]
\caption{Token Lengths of Context}
    \label{table:token_length}
    \centering
    \resizebox{\linewidth}{!}{
    \begin{tabular}{lccc}
    \toprule
         & Instruction & Experience & Policy  \\ \hline
        Twenty Question & 463 & 1011  & 243\\ 
        Who is Undercover & 341 & 2356   & 261  \\ 
        \bottomrule
    \end{tabular}
    }
\end{table}

We aim to qualitatively assess whether current large language models exhibit test-time learning capabilities and the extent to which they improve. To this end, we encode historical experience and compare model performance with and without it.

We investigate efficient and objective methods to encode this historical experience. Table~\ref{table:token_length} reports the average context lengths for instruction, experience, and derived policy. To fully leverage past experience, the experience includes dialogue interactions, rewards, and model's self-reflections on interactions and rewards. The strategy is derived by the model itself based on all past experience.

In pilot studies, we experiment with two approaches: incorporating the full history experience directly, and self-derived policy from the full history. We fix the number of experience to five rounds, leading to context lengths of approximately 5k and 12k for Twenty Questions and Who is Undercover, respectively, while the derived policy contexts average 243 and 261 tokens. Although the first approach provides complete information, it incurs higher computational costs and under-performs compared to the second. Therefore, we adopt policy-based representations of past experience for further evaluation. This setup is illustrated in the left panel of Figure~\ref{Figure:framework}. To further isolate the influence of the model’s self-derived policy pipeline, we include a rule-based policy as a baseline for comparison with the experience-based policy, in which strategies are derived from both rules and accumulated experience. This comparison helps ensure that observed improvements can be attributed to the incorporation of experience.

\subsubsection{Evaluation with Incremental Experience}
\label{sec:cumulative_setting}

The previous setting evaluates the test-time learning given limit amounts of prior experience. If a model demonstrates performance gains from such experience, it becomes essential to investigate whether these test-time improvements persist and accumulate as additional experience is acquired.

This motivates an incremental evaluation setting that requires efficient management of past experience. To support dynamic policy updates with growing experience, we adopt the memory management pipeline~\cite{suzgun2025dynamic}. As illustrated in the right panel of Figure~\ref{Figure:framework}, the agent without experience performs k independent test rounds, while the agent with experience conducts the same k rounds with a continuously updated policy pool based on accumulating experience.
To ensure robust evaluation, we sample each setting (with and without experience) three times and compute the cumulative average reward. Let $r\_{\text{base}}(t, i)$ denote the reward obtained by the agent without experience at test round $t$ in sample $i$, and $r\_{\text{exp}}(t, i)$ denote the corresponding reward for the agent with experience. The cumulative average reward for the agent with experience up to round $t$ is denoted by $R\_{\text{his}}(t)$. 
The computation of $R_{\text{his}}(t)$ is provided below; $R_{\text{base}}(t)$ is computed analogously.

\begin{align}
r_{\text{exp}}(t) &= 
\begin{cases}
\frac{\sum_i r_{\text{base}}(0,i) + \sum_i r_{\text{exp}}(0,i)}{|r_{\text{base}}(0,\cdot)| + |r_{\text{exp}}(0,\cdot)|}, & t = 1 \\
\frac{\sum_i r_{\text{exp}}(t,i)}{|r_{\text{exp}}(t,\cdot)|}, & t > 1
\end{cases} \\
R_{\text{exp}}(t) &= \frac{\sum_{1 \le i \le t}r_{\text{exp}}(i)}{t}
\end{align}

\section{Experiments}
In the experiments, we aim to answer the following questions:\\
(Q1) Do current large language models exhibit the ability to learn at test time?  \\
(Q2) Can large language models achieve stable and consistent improvements when experience accumulates? \\
(Q3) How do humans adapt and improve their performance through experience? \\
(Q4) How do thinking models perform in test-time learning scenarios? \\

\subsection{Experimental Setup}
We aim to evaluate whether the current top-tier large language models have the ability to improve at the test time. Specifically, we evaluate gpt-4o~\cite{hurst2024gpt}, Claude 3.5 Sonnet~\cite{claude_api} and DeepSeek-V3~\cite{liu2024deepseek}. We set the temperature to 1 to support the dynamic testbeds.
For overall performance evaluations, we set prior interactions N=5, test cases M=32, which we find to yield stable results. In the cumulative setting, we extend the evaluation t=50 rounds.

During each interaction, the model is instructed to first perform explicit reasoning before generating its final output. The final response (a question, reflection, or policy in Twenty Questions; a speech, vote, reflection, or policy in Who is Undercover) is enclosed within <answer></answer> tags to ensure clarity and facilitate objective evaluation of both reasoning quality and task performance.

In the single-agent setting, the environment is simulated using the same model under evaluation to ensure alignment in question understanding and knowledge base. In the multi-agent setting, all other agents are instantiated with the same backbone LLM as the test agent to isolate test-time improvements from potential gains due to mere familiarity with another model’s behavior. For each evaluation setting, the order of test rounds is fixed to ensure consistency across trials.

 \begin{table*}[htpb]
\caption{Evaluation of Test-time Learning Ability of LLMs. "w/o Policy" denotes the baseline setting where the model is provided only with task rules. "w/ Rule Policy" indicates that the model receives both the rules and a test-time policy based only on rules. "w/ Exp. Policy" refers to having both rules and test-time policy from rules and model five rounds of experience containing interactions, rewards and reflections. "w/ Human Policy" indicates that the model is given rules along with a human-authored policy based on human understanding of the task. The best results are shown in \textbf{bold} and the second best are \underline{underlined}.}
\label{table:main_result}
\centering
\resizebox{0.9\textwidth}{!}{
\begin{tabular}{llccc}
\toprule
\textbf{Task} & \textbf{Setting} & \textbf{GPT-4o} & \textbf{Claude 3.5 Sonnet} & \textbf{DeepSeek-V3} \\
\midrule
\multirow{3}{*}{\makecell[l]{\textbf{AIME 2025}\\Single-Turn Math Problem}} 
& w/o Policy       & \textbf{0.0792} & \textbf{0.1417} & \textbf{0.2750} \\
& w/\phantom{o} Exp. Policy  & 0.0542 & \textbf{0.1417} & 0.2458 \\
& \textit{\quad Improve (\%)}  & \textit{\ -30.57} &	\textit{\ 0.00} &	\textit{\ -10.62} \\
\midrule
\multirow{7}{*}{\makecell[l]{\textbf{Twenty Question}\\Multi-turn Single-agent}}
& w/o Policy       & 0.2422  & \underline{0.2640} & 0.2641 \\
& w/\phantom{o} Rule Policy  & 0.2199 & 0.1368 & 0.2033 \\
& \textit{\quad Improve (\%)}  & \textit{\ -9.21} &	\textit{\ -48.18}	& \textit{\ -23.02} \\
& w/\phantom{o} Exp. Policy  & \underline{0.2563} & \textbf{0.2807} & \underline{0.2746} \\
& \textit{\quad Improve (\%)}  & \textit{\ 5.80} &	\textit{\ 6.33}	& \textit{\ 3.97} \\
& w/\phantom{o} Human Policy & \textbf{0.2709} & 0.2624  & \textbf{0.2758} \\
& \textit{\quad Improve (\%)}  & \textit{\ 11.84} &	\textit{\ -0.61} & \textit{\ 4.41} \\
\midrule
\multirow{7}{*}{\makecell[l]{\textbf{Who is Undercover}\\Multi-turn Multi-agent}}
& w/o Policy      & 0.1563 & 0.1250 & 0.2500 \\
& w/\phantom{o} Rule Policy  & 0.0625 &  0.3125 &  0.1250 \\
& \textit{\quad Improve (\%)}  & \textit{\ -60.01} &	\textit{\ 150.00} &	\textit{\ -50.00} \\
& w/\phantom{o} Exp. Policy  & \underline{0.1719} & \underline{0.1563} & \underline{0.2813} \\
& \textit{\quad Improve (\%)}  & \textit{\ 10.02} &	\textit{\ 25.04} &	\textit{\ 12.50} \\
& w/\phantom{o} Human Policy & \textbf{0.1875} & \textbf{0.3438} & \textbf{0.4063} \\
& \textit{\quad Improve (\%)}  & \textit{\ 20.00} &	\textit{\ 175.04} &	\textit{\ 62.50}  \\
\bottomrule
\end{tabular}
}
\end{table*}

\subsection{Overall Test-time Learning Performance~(Q1)}

We begin by investigating whether top-performing large language models exhibit measurable improvements at test time when provided with prior experience. Table~\ref{table:main_result} summarizes the overall performance across three environments under four evaluation settings: (1) without any policy, (2) with model-derived policy based solely on rules, (3) with model-derived policy based on both rules and test-time experience, and (4) with human-authored policy. The inclusion of the human policy serves to assess the potentials of models.

In the Twenty Questions setting, we observe consistent performance gains when models are equipped with self-derived policies based on prior experience. In contrast, rule-based policies result in significant performance drops across all models, likely due to a misalignment between human-designed heuristics and model reasoning patterns, as further discussed in Section~\ref{subsec:policy}. Experience-based policies, however, lead to clear improvements, with Claude achieving the highest gain from its own test-time experience.

Interestingly, GPT-4o and DeepSeek-V3 both outperform their self-derived policies when provided with human-authored policies. This highlights a gap between the models' current test-time learning capabilities and their full potential, suggesting that either the quantity of experience or the quality of derived policies remains suboptimal. These limitations are further examined in Section~\ref{subsec:cumulative} and Section~\ref{subsec:policy}. Claude performs marginally worse with human-authored policy, also indicating a possible misalignment between its internal reasoning and externally imposed guidance.

In Who is Undercover, test-time learning yields more substantial improvements. Claude again achieves the highest gain from experience-based policy, reinforcing its ability to leverage self-acquired strategies. Unlike other settings, the rule-based policy ranks as the second-best for some models, highlighting a divergent pattern in this multi-agent context. Additionally, human-authored policies consistently lead to the highest performance across all models, further underscoring the latent potential of test-time learning when guided by effective strategies.
It is important to note that direct comparisons across models in this environment are not meaningful, as all agents in the multi-agent setting are instantiated using the same LLM that is being evaluated. This design ensures an objective assessment of test-time learning by isolating gains attributable to experience and strategic adaptation, rather than confounding effects such as familiarity with another model’s behavior. Full instances of model-generated and human-authored policies are provided in Appendix~\ref{appendix:policy} and analyzed in Section~\ref{subsec:policy}.

\textit{Finding 1: Policies derived from past experience at test time yield measurable improvements across models and tasks.}

\textit{Finding 2: The superior performance under human-authored policies reveals the untapped potential for enhancing models' test-time learning capabilities.}


\subsection{Cumulative Improvement~(Q2)}
\label{subsec:cumulative}

\begin{figure*}[htbp]
  \includegraphics[width=\linewidth]{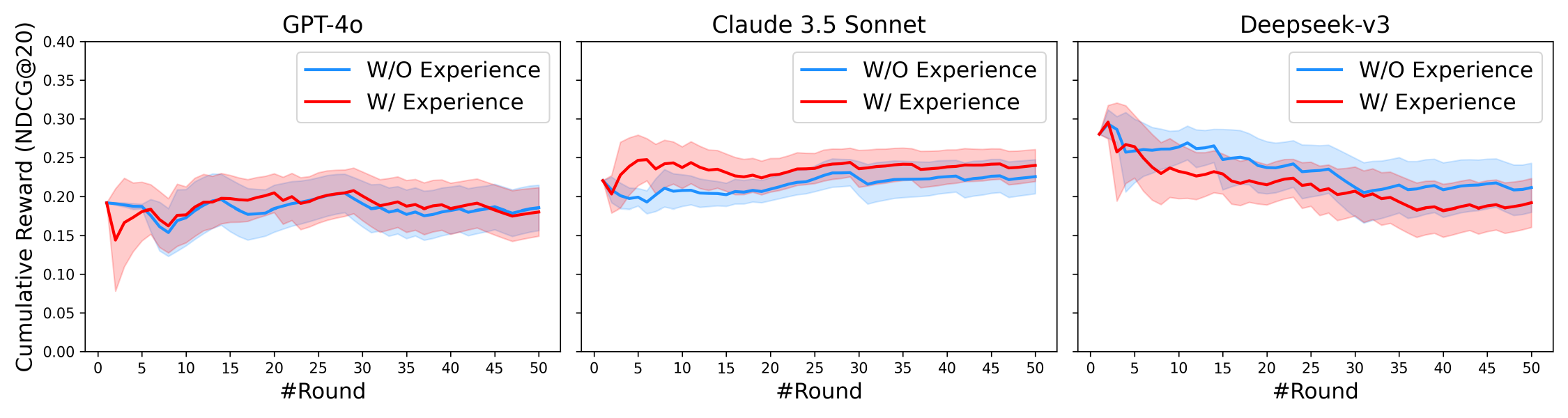} 
  \caption {Cumulative Test-Time Learning Performance on Twenty Question.}
  \label{Figure:cumulative}
\end{figure*}

The above results demonstrate that large language models possess the ability to improve at test time. We next examine whether this improvement is consistent as experience accumulates. To this end, we adopt the cumulative evaluation setting described in Section~\ref{sec:cumulative_setting}. Figure~\ref{Figure:cumulative} presents cumulative rewards over 50 rounds in the Twenty Questions task, comparing model performance with and without test-time policies derived from past experience.

Model performances vary. Claude successfully leverages cumulative experience, whereas other models struggle to maintain or improve performance as experience accumulates. For Claude, the experience-enabled setting consistently outperforms the baseline, particularly within the first five rounds, indicating effective strategy accumulation. However, the performance gap narrows in later rounds, suggesting diminishing returns from additional experience.
GPT and DeepSeek show minimal gains from the accumulation of experience at test time.
For GPT-4o, both curves overlap in the early rounds, with the experience-enabled setting beginning to slightly outperform the baseline around rounds 15–20. In contrast, DeepSeek-V3 shows a decline in performance after five rounds of accumulated experience, while the baseline remains stable. This suggests that its policy refinement process may introduce noise or compounding errors, limiting its ability to leverage experience effectively.


\textit{Finding 3: Results reveal substantial differences in the consistency and effectiveness of test-time learning across models as experience grows.}

\subsection{Human Study~(Q3)}

In the previous experiments, we demonstrate that certain large language models exhibit the ability to learn at test time through cumulative experience. To further understand the rate of improvement, we compare model learning speed with human.

\begin{figure}[htbp]
  \includegraphics[width=0.9\linewidth]{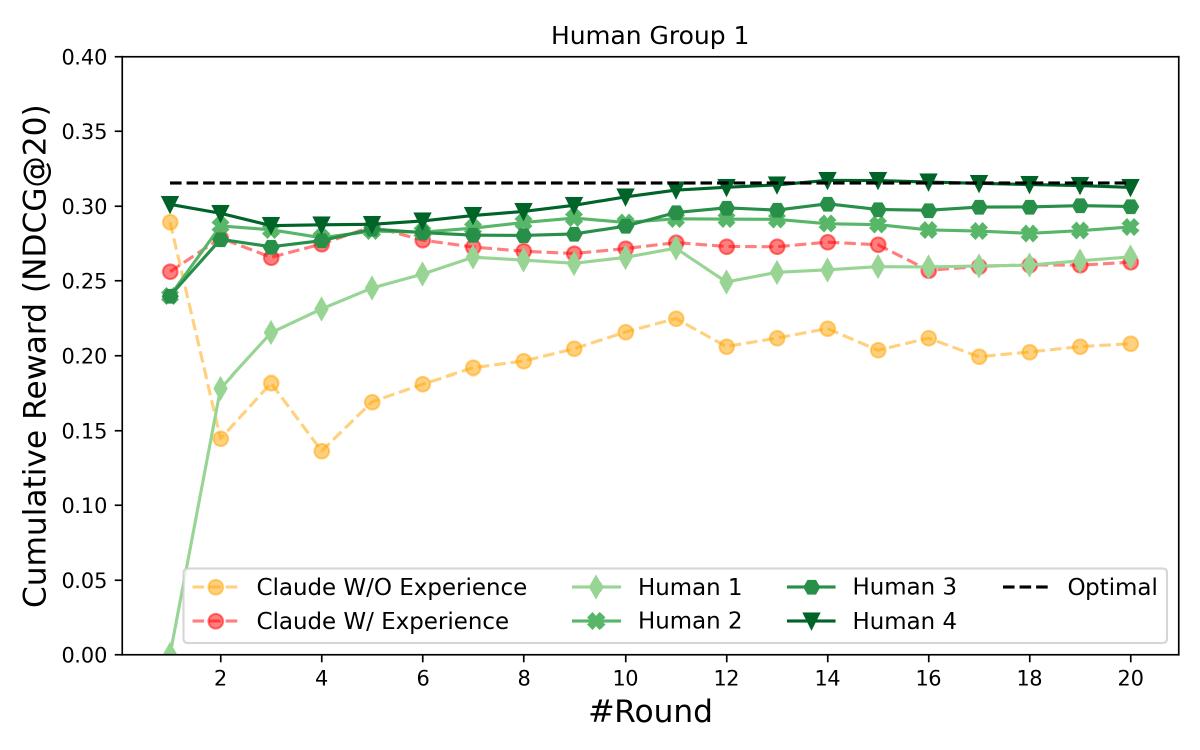} 
  \includegraphics[width=0.9\linewidth]{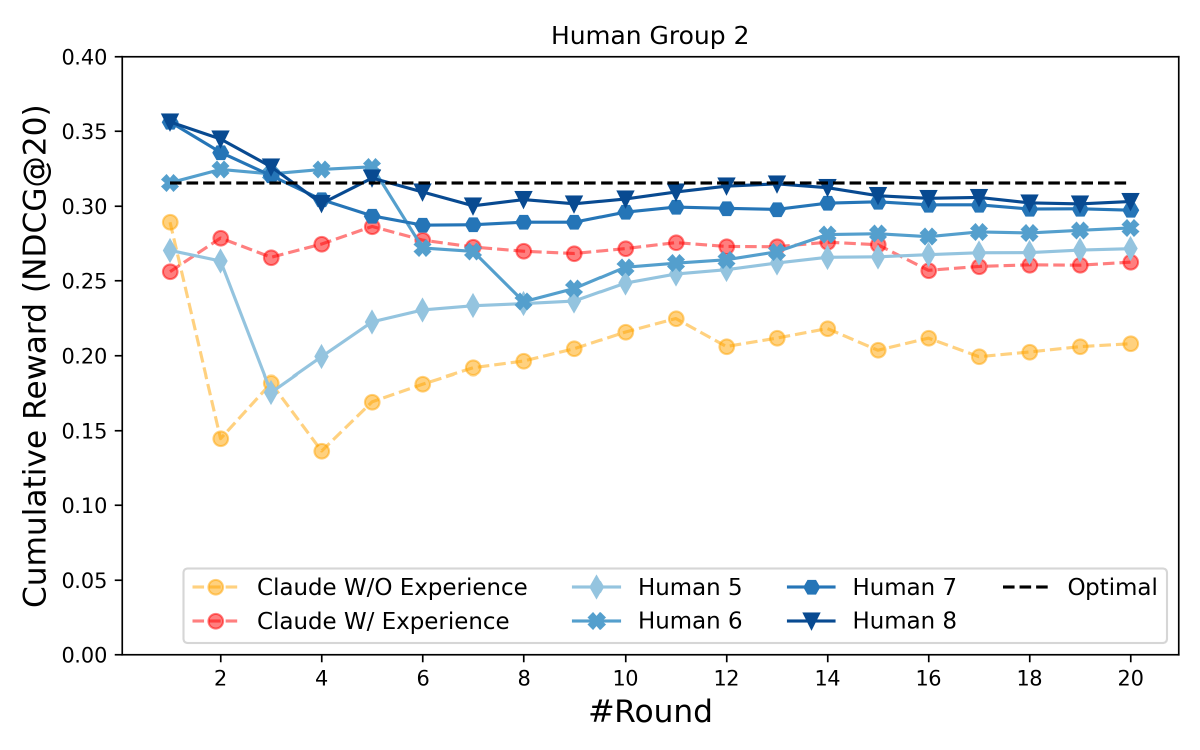} 
  \caption{Human Performance on Twenty Question.}
  \label{Figure:human}
\end{figure}

We recruited eight human participants (undergraduate and PHD students) to perform the same Twenty Question task, playing 20 rounds cumulatively. Their results are summarized in Table~\ref{Figure:human}, and their cumulative rewards are plotted alongside those of the best-performing model, Claude. Participants are divided into two groups based on performance variance across rounds.

The upper figure shows that all humans in this group achieve greater cumulative gains than Claude after 20 rounds, approaching near-optimal performance (represented by the black dotted line indicating the reward of perfect binary questioning). The lower figure includes participants with higher performance variability; nevertheless, their final cumulative rewards still exceed those of the LLM.


\textit{Finding 4: Current top-tier large language models exhibit slower test-time learning speed compared to the learning efficiency of humans in the experience-based reasoning-intensive task.}

\subsection{Performance of Thinking Models~(Q4)}


\begin{table}[htbp]
\caption{Test-time Learning Performance of Thinking Models in Who is Undercover environment.}
    \label{table:thinking_model}
    \centering
    \resizebox{0.95\linewidth}{!}{
    \begin{tabular}{lccc}
    \toprule
       \textbf{Setting}  & \textbf{o1} & \textbf{Claude} & \textbf{R1} \\ \hline
        w/o Test-time Policy        &  \underline{0.30}   & \underline{0.10} &\textbf{0.25}  \\
        w/\phantom{o} Test-time Policy & 0.20 & \underline{0.10} &0.00 \\
        w/\phantom{o} Human Policy & \textbf{0.40} & \textbf{0.20} & \underline{0.20}   \\ \bottomrule
    \end{tabular}
    }
\end{table}

In pervious experiments, we evaluate large language models without explicit thinking mode. In this section, we examine the performance of thinking model: o1~\cite{jaech2024openai} and Deepseek-r1~\cite{guo2025deepseek}, as reported in Table~\ref{table:thinking_model}.

As shown in the table, test-time learning improvements are not observed for either thinking model when provided with self-derived policies.
For o1, the test-time performance with self policy decrease and the test-time performance with human policy increase. 
For o1, performance decreases when incorporate test-time policy but improves when guided by a human policy, suggesting potential limitations in its ability to generate effective strategies autonomously, while still being capable of leveraging prior experience.
For DeepSeek-R1, performance declines under both self-derived and human-authored policy conditions, compared to its baseline with no prior experience. This aligns with the findings reported in its original paper, which notes that few-shot prompting consistently degrades R1’s performance. The authors explicitly recommend presenting tasks in a zero-shot format for optimal outcomes, suggesting that R1’s internal reasoning is optimized for direct problem descriptions rather than for accumulating and adapting to test-time experience.

\textit{Finding 5: Test-time learning is not observed in thinking models, consistent with the findings reported in R1's original paper that CoT in few-shot cases may degrade model performance.}

\subsection{Further Analyses}
\label{subsec:policy}

In the Twenty Questions environment, test-time improvements (\textit{w/ Exp. Policy} vs. \textit{w/o Policy} in Table~\ref{table:main_result}) primarily stem from earlier identification of item categories. Test-time policies such as “Begin with high-level distinctions” and “Identify the category of the answer word within the first five questions” help the model avoid overly specific guesses early on. We also analyze the failure of the \textit{w/ Rule Policy} setting in this environment, which we attribute to a misalignment between model behavior and human preference. Model-generated questions often include specific examples (e.g., “Is it a living thing (animal, plant)?” or “Is it a ball (like basketball, baseball, football) rather than other sports equipment (like bats or rackets)?”), whereas questions in human-authored policies are general and abstract  (e.g., “Is it a living thing?”, “Does it use electricity?”, “Is it commonly found indoors?”). These examples lead to the model adopt this format throughout the questioning process, which contributes to the performance decline.
For humans, we observed more rapid test-time learning, with noticeable improvement after just a single game.


In the Who is Undercover environment, both DeepSeek and human demonstrate a key policy, "Deduce the opposing secret word". This is based on the observation that the undercover and normal players’ words are typically semantically related. Recognizing this pattern allows participants to refine their clues and identities more effectively.

\section{Conclusion}

In this work, we advocate for evaluations of large language models’ test-time learning ability, defined as the capacity to improve at test time in experience-based, reasoning-intensive tasks. Competitive games serve as effective testbeds due to their dynamic nature, resistance to saturation, and reliance on reasoning.
We present an objective framework to access test-time learning under both static and cumulative experience settings and evaluate models with different policies. We also compare the improvements of models to humans.
Experimental results demonstrate that LLMs exhibit measurable test-time learning; however, these gains are often unstable and inconsistent across cumulative settings. In contrast, human participants show more stable and rapid learning. We highlight the need for greater emphasis on evaluation and training strategies to improve LLMs’ test-time learning as a step toward artificial general intelligence.

\section*{Limitations}
This work aims to evaluate the test-time learning capabilities of large language models and compare their gains with humans. We employ one math benchmark and two semantic game environments as representative testbeds. While these settings provide meaningful insights, a broader range of evaluation environments is necessary for more comprehensive measurement of test-time learning.

Each experimental condition is run for thirty two to fifty rounds to ensure result stability. We examine five settings: no experience, rule-based policy without experience, full experience, experience-based policy, and human-authored policy, to provide a more thorough understanding of how LLMs benefit from test-time experience.

For the human studies, we recruit eight participants, each compensated \$10 for completing 20 rounds of the Twenty Questions task, which typically takes 1 to 2 hours. While these participants offer a stable and useful baseline, involving more individuals across diverse environments would further strengthen the generalizability of our findings.

Finally, this paper currently does not incorporate test-time training, where a model's parameters are updated based on a small amount of test-time experience. This is due to two main reasons: the closed-source nature of the models evaluated, which precludes parameter access; and the current lack of effective in-parameter test-time learning methods tailored for reasoning-intensive tasks. Future work may explore this direction to further advance the understanding of test-time learning capabilities in LLMs.

\section*{Ethical Considerations}
In this study, we recruited eight human participants to complete twenty rounds of the Twenty Questions task. The task environment is non-sensitive and does not pose any potential risks or negative impacts to participants. All participants were fully informed that their interactions would be recorded and used for research purposes, and their consent was obtained prior to participation.
In this paper, we used an AI assistant (GPT-4o) to check for grammatical errors. It was not used to directly generate any of the paper's content.

\bibliography{custom}

\begin{thebibliography}{37}
\providecommand{\natexlab}[1]{#1}

\bibitem[{Abdulhai et~al.(2023)Abdulhai, White, Snell, Sun, Hong, Zhai, Xu, and Levine}]{abdulhai2023lmrl}
Marwa Abdulhai, Isadora White, Charlie Snell, Charles Sun, Joey Hong, Yuexiang Zhai, Kelvin Xu, and Sergey Levine. 2023.
\newblock Lmrl gym: Benchmarks for multi-turn reinforcement learning with language models.
\newblock \emph{arXiv preprint arXiv:2311.18232}.

\bibitem[{Achiam et~al.(2023)Achiam, Adler, Agarwal, Ahmad, Akkaya, Aleman, Almeida, Altenschmidt, Altman, Anadkat et~al.}]{achiam2023gpt}
Josh Achiam, Steven Adler, Sandhini Agarwal, Lama Ahmad, Ilge Akkaya, Florencia~Leoni Aleman, Diogo Almeida, Janko Altenschmidt, Sam Altman, Shyamal Anadkat, and 1 others. 2023.
\newblock Gpt-4 technical report.
\newblock \emph{arXiv preprint arXiv:2303.08774}.

\bibitem[{Anthropic(2024)}]{claude_api}
Anthropic. 2024.
\newblock \href {https://www.anthropic.com/news/claude-3-5-sonnet} {Claude 3.5 sonnet}.

\bibitem[{Chang et~al.(2024)Chang, Wang, Wang, Wu, Yang, Zhu, Chen, Yi, Wang, Wang et~al.}]{chang2024survey}
Yupeng Chang, Xu~Wang, Jindong Wang, Yuan Wu, Linyi Yang, Kaijie Zhu, Hao Chen, Xiaoyuan Yi, Cunxiang Wang, Yidong Wang, and 1 others. 2024.
\newblock A survey on evaluation of large language models.
\newblock \emph{ACM transactions on intelligent systems and technology}, 15(3):1--45.

\bibitem[{Chiang et~al.(2023)Chiang, Li, Lin, Sheng, Wu, Zhang, Zheng, Zhuang, Zhuang, Gonzalez, Stoica, and Xing}]{vicuna2023}
Wei-Lin Chiang, Zhuohan Li, Zi~Lin, Ying Sheng, Zhanghao Wu, Hao Zhang, Lianmin Zheng, Siyuan Zhuang, Yonghao Zhuang, Joseph~E. Gonzalez, Ion Stoica, and Eric~P. Xing. 2023.
\newblock \href {https://lmsys.org/blog/2023-03-30-vicuna/} {Vicuna: An open-source chatbot impressing gpt-4 with 90\%* chatgpt quality}.

\bibitem[{Cobbe et~al.(2021)Cobbe, Kosaraju, Bavarian, Chen, Jun, Kaiser, Plappert, Tworek, Hilton, Nakano, Hesse, and Schulman}]{cobbe2021gsm8k}
Karl Cobbe, Vineet Kosaraju, Mohammad Bavarian, Mark Chen, Heewoo Jun, Lukasz Kaiser, Matthias Plappert, Jerry Tworek, Jacob Hilton, Reiichiro Nakano, Christopher Hesse, and John Schulman. 2021.
\newblock Training verifiers to solve math word problems.
\newblock \emph{arXiv preprint arXiv:2110.14168}.

\bibitem[{Duan et~al.(2016)Duan, Schulman, Chen, Bartlett, Sutskever, and Abbeel}]{duan2016rl}
Yan Duan, John Schulman, Xi~Chen, Peter~L Bartlett, Ilya Sutskever, and Pieter Abbeel. 2016.
\newblock Rl2: Fast reinforcement learning via slow reinforcement learning.
\newblock \emph{arXiv preprint arXiv:1611.02779}.

\bibitem[{Gandelsman et~al.(2022)Gandelsman, Sun, Chen, and Efros}]{gandelsman2022test}
Yossi Gandelsman, Yu~Sun, Xinlei Chen, and Alexei Efros. 2022.
\newblock Test-time training with masked autoencoders.
\newblock \emph{Advances in Neural Information Processing Systems}, 35:29374--29385.

\bibitem[{Garivier and Moulines(2011)}]{garivier2011upper}
Aur{\'e}lien Garivier and Eric Moulines. 2011.
\newblock On upper-confidence bound policies for switching bandit problems.
\newblock In \emph{International conference on algorithmic learning theory}, pages 174--188. Springer.

\bibitem[{Grigsby et~al.()Grigsby, Fan, and Zhu}]{grigsby2023amago}
Jake Grigsby, Linxi Fan, and Yuke Zhu.
\newblock Amago: Scalable in-context reinforcement learning for adaptive agents.
\newblock In \emph{The Twelfth International Conference on Learning Representations}.

\bibitem[{Guo et~al.(2025)Guo, Yang, Zhang, Song, Zhang, Xu, Zhu, Ma, Wang, Bi et~al.}]{guo2025deepseek}
Daya Guo, Dejian Yang, Haowei Zhang, Junxiao Song, Ruoyu Zhang, Runxin Xu, Qihao Zhu, Shirong Ma, Peiyi Wang, Xiao Bi, and 1 others. 2025.
\newblock Deepseek-r1: Incentivizing reasoning capability in llms via reinforcement learning.
\newblock \emph{arXiv preprint arXiv:2501.12948}.

\bibitem[{Hurst et~al.(2024)Hurst, Lerer, Goucher, Perelman, Ramesh, Clark, Ostrow, Welihinda, Hayes, Radford et~al.}]{hurst2024gpt}
Aaron Hurst, Adam Lerer, Adam~P Goucher, Adam Perelman, Aditya Ramesh, Aidan Clark, AJ~Ostrow, Akila Welihinda, Alan Hayes, Alec Radford, and 1 others. 2024.
\newblock Gpt-4o system card.
\newblock \emph{arXiv preprint arXiv:2410.21276}.

\bibitem[{Jaech et~al.(2024)Jaech, Kalai, Lerer, Richardson, El-Kishky, Low, Helyar, Madry, Beutel, Carney et~al.}]{jaech2024openai}
Aaron Jaech, Adam Kalai, Adam Lerer, Adam Richardson, Ahmed El-Kishky, Aiden Low, Alec Helyar, Aleksander Madry, Alex Beutel, Alex Carney, and 1 others. 2024.
\newblock Openai o1 system card.
\newblock \emph{arXiv preprint arXiv:2412.16720}.

\bibitem[{Jiang et~al.(2023)Jiang, Wang, and Wang}]{jiang2023selfevolve}
Shuyang Jiang, Yuhao Wang, and Yu~Wang. 2023.
\newblock Selfevolve: A code evolution framework via large language models.
\newblock \emph{arXiv preprint arXiv:2306.02907}.

\bibitem[{Krathwohl(2002)}]{krathwohl2002revision}
David~R Krathwohl. 2002.
\newblock A revision of bloom's taxonomy: An overview.
\newblock \emph{Theory into practice}, 41(4):212--218.

\bibitem[{Lange et~al.(2024)Lange, Tian, and Tang}]{lange2024large}
Robert Lange, Yingtao Tian, and Yujin Tang. 2024.
\newblock Large language models as evolution strategies.
\newblock In \emph{Proceedings of the Genetic and Evolutionary Computation Conference Companion}, pages 579--582.

\bibitem[{Laskin et~al.()Laskin, Wang, Oh, Parisotto, Spencer, Steigerwald, Strouse, Hansen, Filos, Brooks et~al.}]{laskin2022context}
Michael Laskin, Luyu Wang, Junhyuk Oh, Emilio Parisotto, Stephen Spencer, Richie Steigerwald, DJ~Strouse, Steven~Stenberg Hansen, Angelos Filos, Ethan Brooks, and 1 others.
\newblock In-context reinforcement learning with algorithm distillation.
\newblock In \emph{The Eleventh International Conference on Learning Representations}.

\bibitem[{Lee et~al.(2023)Lee, Xie, Pacchiano, Chandak, Finn, Nachum, and Brunskill}]{lee2023supervised}
Jonathan Lee, Annie Xie, Aldo Pacchiano, Yash Chandak, Chelsea Finn, Ofir Nachum, and Emma Brunskill. 2023.
\newblock Supervised pretraining can learn in-context reinforcement learning.
\newblock \emph{Advances in Neural Information Processing Systems}, 36:43057--43083.

\bibitem[{Liu et~al.(2024)Liu, Feng, Xue, Wang, Wu, Lu, Zhao, Deng, Zhang, Ruan et~al.}]{liu2024deepseek}
Aixin Liu, Bei Feng, Bing Xue, Bingxuan Wang, Bochao Wu, Chengda Lu, Chenggang Zhao, Chengqi Deng, Chenyu Zhang, Chong Ruan, and 1 others. 2024.
\newblock Deepseek-v3 technical report.
\newblock \emph{arXiv preprint arXiv:2412.19437}.

\bibitem[{Liu et~al.(2021)Liu, Kothari, Van~Delft, Bellot-Gurlet, Mordan, and Alahi}]{liu2021ttt++}
Yuejiang Liu, Parth Kothari, Bastien Van~Delft, Baptiste Bellot-Gurlet, Taylor Mordan, and Alexandre Alahi. 2021.
\newblock Ttt++: When does self-supervised test-time training fail or thrive?
\newblock \emph{Advances in Neural Information Processing Systems}, 34:21808--21820.

\bibitem[{Lu et~al.(2023{\natexlab{a}})Lu, Schroecker, Gu, Parisotto, Foerster, Singh, and Behbahani}]{lu2023structured}
Chris Lu, Yannick Schroecker, Albert Gu, Emilio Parisotto, Jakob Foerster, Satinder Singh, and Feryal Behbahani. 2023{\natexlab{a}}.
\newblock Structured state space models for in-context reinforcement learning.
\newblock \emph{Advances in Neural Information Processing Systems}, 36:47016--47031.

\bibitem[{Lu et~al.(2023{\natexlab{b}})Lu, Zhong, Huang, Wang, Mi, Wang, Wang, Shang, and Liu}]{lu2023self}
Jianqiao Lu, Wanjun Zhong, Wenyong Huang, Yufei Wang, Fei Mi, Baojun Wang, Weichao Wang, Lifeng Shang, and Qun Liu. 2023{\natexlab{b}}.
\newblock Self: Language-driven self-evolution for large language model.

\bibitem[{Luo et~al.(2023)Luo, Xu, Zhao, Sun, Geng, Hu, Tao, Ma, Lin, and Jiang}]{luo2023wizardcoder}
Ziyang Luo, Can Xu, Pu~Zhao, Qingfeng Sun, Xiubo Geng, Wenxiang Hu, Chongyang Tao, Jing Ma, Qingwei Lin, and Daxin Jiang. 2023.
\newblock Wizardcoder: Empowering code large language models with evol-instruct.
\newblock \emph{arXiv preprint arXiv:2306.08568}.

\bibitem[{MAA(2025)}]{AIME}
MAA. 2025.
\newblock \href {https://maa.org/} {American invitational mathematics examination (aime)}.

\bibitem[{Minsky(1988)}]{minsky1988society}
Marvin Minsky. 1988.
\newblock \emph{Society of mind}.
\newblock Simon and Schuster.

\bibitem[{Morad et~al.(2023)Morad, Kortvelesy, Bettini, Liwicki, and Prorok}]{morad2023popgym}
Steven Morad, Ryan Kortvelesy, Matteo Bettini, Stephan Liwicki, and Amanda Prorok. 2023.
\newblock Popgym: Benchmarking partially observable reinforcement learning.
\newblock \emph{arXiv preprint arXiv:2303.01859}.

\bibitem[{Qian et~al.(2024)Qian, Liang, Qin, Ye, Cong, Lin, Wu, Liu, and Sun}]{qian2024investigate}
Cheng Qian, Shihao Liang, Yujia Qin, Yining Ye, Xin Cong, Yankai Lin, Yesai Wu, Zhiyuan Liu, and Maosong Sun. 2024.
\newblock Investigate-consolidate-exploit: A general strategy for inter-task agent self-evolution.
\newblock \emph{arXiv preprint arXiv:2401.13996}.

\bibitem[{Silver and Sutton(2025)}]{silver2025welcome}
David Silver and Richard~S Sutton. 2025.
\newblock Welcome to the era of experience.
\newblock \emph{Google AI}.

\bibitem[{Sinha et~al.(2023)Sinha, Gehler, Locatello, and Schiele}]{sinha2023test}
Samarth Sinha, Peter Gehler, Francesco Locatello, and Bernt Schiele. 2023.
\newblock Test: Test-time self-training under distribution shift.
\newblock In \emph{Proceedings of the IEEE/CVF Winter Conference on Applications of Computer Vision}, pages 2759--2769.

\bibitem[{Sun et~al.(2020)Sun, Wang, Liu, Miller, Efros, and Hardt}]{sun2020test}
Yu~Sun, Xiaolong Wang, Zhuang Liu, John Miller, Alexei Efros, and Moritz Hardt. 2020.
\newblock Test-time training with self-supervision for generalization under distribution shifts.
\newblock In \emph{International conference on machine learning}, pages 9229--9248. PMLR.

\bibitem[{Sun et~al.(2019)Sun, Wang, Liu, Miller, Efros, and Hardt}]{sun2019test}
Yu~Sun, Xiaolong Wang, Zhuang Liu, John Miller, Alexei~A Efros, and Moritz Hardt. 2019.
\newblock Test-time training for out-of-distribution generalization.

\bibitem[{Suzgun et~al.(2025)Suzgun, Yuksekgonul, Bianchi, Jurafsky, and Zou}]{suzgun2025dynamic}
Mirac Suzgun, Mert Yuksekgonul, Federico Bianchi, Dan Jurafsky, and James Zou. 2025.
\newblock Dynamic cheatsheet: Test-time learning with adaptive memory.
\newblock \emph{arXiv preprint arXiv:2504.07952}.

\bibitem[{Tao et~al.(2024)Tao, Lin, Chen, Li, Wu, Li, Jin, Huang, Tao, and Zhou}]{tao2024survey}
Zhengwei Tao, Ting-En Lin, Xiancai Chen, Hangyu Li, Yuchuan Wu, Yongbin Li, Zhi Jin, Fei Huang, Dacheng Tao, and Jingren Zhou. 2024.
\newblock A survey on self-evolution of large language models.
\newblock \emph{arXiv preprint arXiv:2404.14387}.

\bibitem[{Xu et~al.(2023)Xu, Hu, Zhou, Ren, Dong, Keutzer, Ng, and Feng}]{xu2023magic}
Lin Xu, Zhiyuan Hu, Daquan Zhou, Hongyu Ren, Zhen Dong, Kurt Keutzer, See~Kiong Ng, and Jiashi Feng. 2023.
\newblock Magic: Investigation of large language model powered multi-agent in cognition, adaptability, rationality and collaboration.
\newblock \emph{arXiv preprint arXiv:2311.08562}.

\bibitem[{Yao et~al.(2022)Yao, Chen, Yang, and Narasimhan}]{yao2022webshop}
Shunyu Yao, Howard Chen, John Yang, and Karthik Narasimhan. 2022.
\newblock Webshop: Towards scalable real-world web interaction with grounded language agents.
\newblock \emph{Advances in Neural Information Processing Systems}, 35:20744--20757.

\bibitem[{Yu and Feng(2025)}]{yu2025policyevol}
Yajie Yu and Yue Feng. 2025.
\newblock Policyevol-agent: Evolving policy via environment perception and self-awareness with theory of mind.
\newblock \emph{arXiv preprint arXiv:2504.15313}.

\bibitem[{Zhou et~al.(2024)Zhou, Zanette, Pan, Levine, and Kumar}]{zhou2024archer}
Yifei Zhou, Andrea Zanette, Jiayi Pan, Sergey Levine, and Aviral Kumar. 2024.
\newblock Archer: Training language model agents via hierarchical multi-turn rl.
\newblock \emph{arXiv preprint arXiv:2402.19446}.

\end{thebibliography}

\appendix

\section*{Appendix}
\label{sec:appendix}

\section{Game Rule}
\subsection{Twenty Question}

You are playing the game Twenty Questions. \\
In this game, there are 157 candidate words: Airplane, Apple, Banana, Baseball, Baseball bat, Basketball, Battery, Bear, Bed, Belt, Blender, Boat, Bookcase, Boots, Bowl, Bracelet, Broccoli, Brooch, Bus, Bush, Cactus, Calculator, Calendar, Camera, Cantaloupe, Canvas, Car, Carrot, Cat, Celery, Chair, Chopstick, Clarinet, Computer, Computer keyboard, Cooking pot, Corn, Couch, Cow, Cucumber, Cup, Desk, Diary, Dog, Doll, Dress, Dresser, Drill, Drum, Earring, Elephant, Eraser, Flute, Football, Forest, Fork, Gloves, Glue, Golf ball, Grape, Guitar, Hairclip, Hammer, Harp, Hat, Headphone, Helicopter, Helmet, Horse, Jacket, Key, Kite, Knife, Lake, Lawn mower, Lego, Lion, Locket, Mango, Marker, Mattress, Meteorite, Microwave, Monitor, Motorcycle, Mountain, Necklace, Nightstand, Ocean, Onion, Orange, Paintbrush, Painting, Pan, Pants, Paper, Peach, Peas, Pen, Pencil, Pendant, Piano, Pillow, Pineapple, Plate, Pliers, Potato, Printer, Puzzle, Rabbit, Rake, Refrigerator, Ring, River, Rock, Saw, Saxophone, Scarf, Scissors, Scooter, Screwdriver, Sculpture, Sea, Sharpie, Sheep, Ship, Shirt, Shoes, Shovel, Skirt, Smartphone, Soccer ball, Socks, Spinach, Spoon, Stapler, Strawberry, Table, Television, Tennis ball, Tennis racket, Tiger, Tomato, Toothbrush, Toothpaste, Train, Tree, Trombone, Truck, Trumpet, Violin, Volleyball, Watch, Watering can, Watermelon, Whisk, Wrench.\\
One of the 157 words is the answer word. You can ask up to 20 yes/no questions to identify the answer word. You will get a Yes, No, or Invalid feedback for each question. Note that you cannot ask questions about the letters of the answer word. Questions can only be asked around semantics.

\subsection{Who is Undercover}

Game Rules of Who is the Difference

1. Game Setup\\
Roles:\\
Among all players, only one player is the difference one. All others are normal players.\\
Secret Words:\\
All normal players receive the same secret word. The difference receives a related but different word.
Initial Information:\\
Each player knows only their own word, not their role or others' words.\\

2. Game Phases\\
The game starts with three rounds of speaking phase, followed by one round of voting. After that, the game alternates between one round of speaking and one round of voting until it ends.
Game ends when the difference is voted out, or the number of remaining players drops to three or fewer.\\
Speaking Phase\\
All players speak in turn, each giving one sentence as a clue about their word. \\
Forbidden actions include saying the word directly, hinting at letters, repeating clues from previous players.
You do not have to describe your own word. The difference may pretend to match the regular players' word to avoid detection.\\
Voting Phase\\
All players vote for one player (cannot be yourself) at the same time. Regular players vote for who they think is the difference; the difference votes strategically.\\
The player with the most votes is eliminated. In case of a tie, no one is eliminated that round.\\
Victory Conditions\\
If the difference is voted out: regular players win \\
If the difference survives until only three players remain: difference wins\\

3. Strategy Tips\\
1. Identify your role early:\\
If more than one player’s clue doesn’t match your word, you are likely the difference.
If only one clue differs, you may be a regular player or the difference. Do not jump to conclusions.

\section{Learnt Policy}
\label{appendix:policy}
\subsection{Twenty Question}

\begin{tcolorbox}[title = {GPT Test-time Policy from Experience},breakable]
\small
To consistently improve performance in Twenty Questions, begin by asking broad, high-level questions that eliminate major categories (e.g., "Is it living?" or "Is it commonly found inside a house?"). Gradually refine with mid-game questions that target functional properties (purpose, interaction, context) rather than overly specific traits. Avoid semantic overload by keeping questions clear, concise, and focused on one characteristic at a time. If progress stalls, pivot to regroup broader ideas rather than fixating on narrow assumptions. Ultimately, efficient grouping and elimination of related possibilities will maximize the chances of identifying the answer within 20 questions.

\end{tcolorbox}



\begin{tcolorbox}[title = {Claude Test-time Policy from Experience},breakable]
\small
\textbf{1. Core Question Structure}\\
- Always start with ``Is it a man-made object?'' \\
- Follow with major category splits like: \\
\quad $\ast$ Electronic vs. non-electronic \\
\quad $\ast$ Indoor vs. outdoor use \\
\quad $\ast$ Portable vs. stationary \\
- Use clear binary choices with ``rather than'' comparisons to avoid invalid responses \\

\textbf{2. Category Hierarchy (in order)}\\
A. Man-made vs. Natural \\
B. If man-made: \\
\quad - Electronic/Electrical vs. Non-electronic \\
\quad - Indoor vs. Outdoor use \\
\quad - Major categories: \\
\qquad $\ast$ Electronics/Devices \\
\qquad $\ast$ Furniture \\
\qquad $\ast$ Clothing/Accessories \\
\qquad $\ast$ Tools/Equipment \\
\qquad $\ast$ Sports/Recreation \\
\qquad $\ast$ Kitchen/Household items \\

\textbf{3. Question Formulation Rules}\\
\quad - Use clear yes/no format \\
\quad - Avoid questions about letters or spelling \\
\quad - Include comparison elements (e.g., ``rather than'') \\
\quad - Focus on physical characteristics and use cases \\
\quad - Avoid ambiguous or overlapping categories \\

\textbf{4. Strategic Principles}\\
\quad - Each question should aim to eliminate roughly half of remaining possibilities \\
\quad - Track eliminated categories carefully \\
\quad - Move from general to specific progressively \\
\quad - Don't get too specific too early \\
\quad - Save direct object identification for final questions \\
\end{tcolorbox}

\begin{tcolorbox}[title = {DeepSeek Test-time Policy from Experience},breakable]
\small
1. \textbf{Start Broad, Then Narrow}: 

\quad Begin with high-level categories (living/non-living, household/outdoor) to eliminate large groups quickly. Prioritize binary splits to maximize information gain early.  

2. \textbf{Focus on Function and Context}: 

\quad For household items, drill down by room (kitchen/bathroom) or usage (cleaning, recreation, grooming). Ask about *how* or *where* the item is used (e.g., "Is it for cutting?" or "Found in the bathroom?").  

3. \textbf{Avoid Invalid Questions}: 

\quad  Use clear, single-focus questions (e.g., "Is it jewelry?" not "Is it worn on hands or neck?"). Split compound queries to prevent ambiguity.  

4. \textbf{Guess Strategically in Small Pools}: 

\quad  Once candidates are under ~5, guess high-probability words (e.g., "spoon" after "kitchen utensil") instead of further elimination.  

5. \textbf{Adapt and Pivot}:

\quad If a category yields "No," immediately switch to the next most likely group (e.g., from "tools" to "decorative items"). Leverage feedback (e.g., "Invalid") to refine questioning.  

\textbf{Key Principle}: 

\quad Each question should halve the remaining candidates, balancing systematic elimination with timely guessing.
\end{tcolorbox}

\begin{tcolorbox}[title = {Human Policy},breakable]
\small



1. \textbf{Start Board}: Begin with high-level distinctions to quickly eliminate large groups of items. Useful splits include:\\
   - Is it a living thing?\\
   - Does it use electoricity?\\
   - Is it commonly found indoors?
   
2. \textbf{Identify the category early}: Aim to determine the category of the answer word within the first 5 questions. Avoid overly specific guesses at this stage.\\
	- Major categories among the 157 words include: fruit, vegetable, animal, transportation, sports equipment, musical instrument, electronic product, office equipment, furniture, kitchenware (for food preparing, eating), tools, entertainment items, clothing, and accessories, daily necessity, etc.
    
3. \textbf{Refine with category traits}: Once the category is identified, filter the candidate words accordingly. Then continue binary questioning using finer distinctions within that group.

4. \textbf{Guess when narrowed down}: If only 3 or fewer candidates remain, direct guessing becomes viable. Otherwise, continue using binary splits for efficiency.

5. \textbf{Your question should always be clear}: Avoid compound questions that can lead to invalid responses.
\end{tcolorbox}

\subsection{Who is Undercover}

\begin{tcolorbox}[title = {GPT test-time Policy},breakable]
\small
1. \textbf{Identify Your Role as Early as Possible}: If the descriptions given by more than one player do not match your secret word, then you are the "different one." If only one player's statements conflict with yours, you might be an ordinary player or the "different one," but you cannot determine this immediately.

2. \textbf{Make Your Statements Ambiguous to Avoid Revealing Core Characteristics}: Regardless of whether you are an ordinary player or the "different one," try to use broad and ambiguous descriptions during the game. Avoid directly mentioning distinctive or obvious characteristics of the secret word to reduce the risk of being suspected.

3. \textbf{Spot Deviations and Pinpoint Suspicious Targets}: Over multiple rounds of statements, focus on characteristics that significantly deviate from the majority of players. Identify semantic inconsistencies. The "different one" usually struggles to fully match the clues provided by ordinary players, which may reveal their identity. Use these deviations to precisely target your voting.

4. \textbf{Skillfully Imitate and Blend Into Group Clues}: If you are the "different one," observe the speaking style of the ordinary players and provide vague but relevant supplementary clues based on their descriptions. This will help align yourself with the group and make your statements appear "reasonable" and "credible," minimizing suspicion.

5. \textbf{Adapt Dynamically and Respond Flexibly}: Adjust your speaking strategy according to changes in the game's dynamics. If your statements are overly suspected, try to further obscure your clues and emphasize common characteristics to divert suspicion.

\end{tcolorbox}

\begin{tcolorbox}[title = {Claude Test-time Policy},breakable]
\small
1. \textbf{Identify Identity Early}: If more than one player's description does not match your secret word, then you are likely the different one. If only one player's statement contradicts yours, you could be an ordinary player or the different one, and you cannot make an immediate judgment.

2. \textbf{Pay Attention to Consistency and Subtle Differences}: During the game, closely observe the statements of other players and note their subtle differences. If a player's description is significantly inconsistent with the majority, it may indicate they are the different one.

3. \textbf{Use Vague Descriptions}: Try to provide vague descriptions and avoid being too specific about your secret word. Choose characteristics that can apply to multiple items.

4. \textbf{Strategic Voting}: In critical voting rounds, do not always follow the majority. Sometimes strategically scattering votes can help you achieve your victory conditions, especially if you are the different one.

5. \textbf{Adjust Identity Judgments Timely}: Adjust your identity judgment based on other players' statements. If you find that other players' descriptions do not match your secret word, consider that you might be the different one and adjust your speaking strategy accordingly to avoid revealing your identity.

6. \textbf{Provide Unique Clues}: When speaking, try to provide unique clues that differ from those of other players. Avoid repeating characteristics that have already been mentioned to prevent exposing your secret word.
\end{tcolorbox}

\begin{tcolorbox}[title = {DeepSeek Test-time Policy},breakable]
\small
1. \textbf{Identify your role as early as possible}: If more than one player's description does not match your secret word, then you are the differentiator. If only one player's statement contradicts yours, you could be either a regular player or the differentiator, and cannot determine your role immediately.  

2. \textbf{Deduce the opposing secret word}: Regular players and differentiators have different secret words, but there is a certain connection between them. Both regular players and differentiators should deduce the opposing secret word based on their own word and previous players' statements as quickly as possible.  

3. \textbf{Speaking strategy}: When uncertain about your role, provide relevant but broad clues to avoid statements that directly reveal your secret word. Ideally, these clues should align with the main characteristics of both your secret word and the opposing secret word. Once your role is confirmed, the differentiator should immediately begin describing the regular players' secret word while mimicking their speaking style to avoid suspicion. Regular players, on the other hand, should identify and highlight the most divergent player to pinpoint the sole differentiator.

\end{tcolorbox}

\begin{tcolorbox}[title = {Human Policy},breakable]
\small
1. \textbf{Analyze comprehensively}: other people's statements may apply not only to their own secret word, but to other words as well, and they need to be analyzed and stated in a way that takes into account the scope of application of the description.

2. \textbf{Guessing and targeting the opponent's secret word}: The secret words of normal players and different people are related, but different. It is possible to guess the opponent's secret word for your own secret word, and to narrow down and target the opponent's secret word after someone else has spoken.

3. \textbf{Identify yourself as early as possible}: If >1 player's description does not match your own secret word, then you are a different person. If only 1 player's statement currently contradicts your own, you may be either a regular player or a dissenter and cannot be immediately recognized.

4. \textbf{Speaking with caution. }
In the early stages of the game, when you cannot identify yourself, you need to speak very carefully. Do not reveal the unique clues to your secret word in your speech; choose a description that is broader and applies to a class of words (including your own and your opponent's secret word) to prevent being too specific and revealing your secret word in the first place, as well as avoiding speeches that might cause conflict. \\
Later in the game, after clarifying their identity, the regular player continues to describe their secret word and the Dissenter describes the regular player's secret word. Statements can be chosen for the popular attributes of the words to be described, avoiding describing overly niche clues that could trigger concern and suspicion.
\end{tcolorbox}

\section{Value Alignment}
\label{section:value}

In Table~\ref{tab:value_alignment}, we present that the importance to use neutral expressions as the role names with negative connotations can create confusion and lead to denial when agent are assessing their identity.

\begin{table*}[htbp]
\caption{Deepseek-r1's thinking content about its gaming role identity. Using role names with negative connotations can create confusion and lead to denial when assessing whether it is "undercover". By using neutral role names, it can quickly and directly recognize its identity.}
\label{tab:value_alignment}
    \centering
    \begin{tabular}{p{1.6cm}|p{11cm}}
    \toprule
    Role Name  & Thinking Context \\ \hline
    Undercover  & Based on the above analysis, the other four agents' statements are all more on the side of soymilk, and my word is milk, \textbf{so maybe I'm the undercover agent?} But according to the rules, there is only one undercover agent, while all the other civilians should be milk, but apparently their descriptions are more consistent with soy milk. \textbf{This suggests that I might be the undercover agent}, and it seems contradictory that my secret word is actually soymilk, but the system tells me my word is milk. \textbf{\underline{Or maybe there are other players who are undercover}} and their descriptions are misleading. But according to the game settings, the word for civilians is the same and the word for undercover agents is different but related. If my word is milk, and the other civilians' word should be milk, but they speak like soy milk, \textbf{\underline{that means they might all be undercover?}} That can't be true because there is only one undercover agent. So maybe my judgment is wrong, \textbf{\underline{or the undercover agent is among the others}} and their descriptions are trying to confuse ...... \textbf{\underline{So as a civilian, I should point out their anomalies.}} \\ \hline
    Difference    &  Taken together, the descriptions of the other players seem to point to soymilk, whereas my secret word is milk, \textbf{so I may be the Different One}. As the Different One, I need to make the statement fit the secret word of the regular player (i.e. soy milk). \\
    \bottomrule
    \end{tabular}
\end{table*}

\end{document}